# Hierarchical Gated Recurrent Neural Tensor Network for Answer Triggering

Wei Li    Yunfang Wu[(✉)]

Key Laboratory of Computational Linguistics (Peking University), Ministry of Education
School of Electronic Engineering and Computer Science, Peking University
liweitj47@pku.edu.cn wuyf@pku.edu.cn

**Abstract.** In this paper, we focus on the problem of answer triggering addressed by Yang et al. (2015), which is a critical component for a real-world question answering system. We employ a hierarchical gated recurrent neural tensor (HGRNT) model to capture both the context information and the deep interactions between the candidate answers and the question. Our result on F value achieves 42.6%, which surpasses the baseline by over 10 %.

## 1    Introduction

Answer triggering is a crucial subtask of the open domain question answering (QA) system. It is first brought up by Yang et al. (2015), where the goal is first to detect whether there exist answers in a set of candidate sentences for a question, and if so return the correct answer. This problem is similar to answer selection (AS) in the way that they all include selecting sentence(s) out of a paragraph. The difference is that AS tasks guarantee that there is at least one answer. Trec-QA (Wang et al., 2007) and WikiQA (Yang et al., 2015) have been the benchmark for such problems.

However, the assumption that at least one answer can be found in the candidate sentences may not be true for real-world applications. In many cases, none of the candidate sentences in the retrieved paragraph can answer the question. As reported by Yang et al. (2015), about 2/3 of the questions don't have any correct answers in the related paragraph in the WikiQA dataset. Therefore they claim that answer triggering task is essential in a real-world QA system. Unfortunately, most of the previous researchers neglect this problem and only concentrate on those questions that have correct answers. They either get rid of the unanswerable questions during the data construction procedure (Wang et al., 2007) or omit the unanswerable questions directly when predicting, for instance, Wang and Jiang (2016); Wang et al. (2016, 2017).

Although recent works that focus on measuring the similarity between an individual candidate answer and its corresponding question have reached very good MRR and MAP scores, they ignore the fact that these candidate answer sentences are continuous text in a paragraph in the setting of WikiQA. These sentences are not separate frag-ments, but under a common topic. Based on this observation, we assume that by bringing the context information of the sentences into consideration, we can get better results in the answer triggering problem. This assumption is verified by our experiments. The F score reaches 42.6% in the answer triggering problem of WikiQA, which surpasses the baseline in Yang et al. (2015) by 10%.

Our contributions lie in the following two aspects:

1. We bring attention to the problem of answer triggering, which is very important but has not been thoroughly studied. We improve the F score by 10% over the original baseline model.
2. We employ a hierarchical recurrent neural tensor (HGRNT) model to take context information into consideration when predicting whether a sentence is a correct answer towards the question. Our experiments demonstrate that the context information consistently increases the F score no matter what sentence encoder structures are used.

## 2    Related Work

In the previous studies, researchers tend to focus on the ranking part of the answer selection (AS) problem, what they need to do is to extract the most probable one from a set of pre-selected sentences. Traditional approaches calculate the similarity of two sentences based on hand crafted features (Yao et al., 2013; Heilman and Smith, 2010; Severyn and Moschitti, 2013). As deep learning thrives, researchers turn to deep learning methods. At the early stage, they apply neural networks like recurrent neural networks (RNN) or convolutional neural networks (CNN) to encode each of the sentences into a fixed length vector, and then compare the question and answer by calculating the semantic distance between these two vectors (Feng et al., 2015; Wang and Nyberg, 2015).

Recent works focus on bringing attention mechanism into the question answering problem inspired by the success of attention based machine translation (Bahdanau et al., 2014). Hermann et al. (2015) and Tan et al. (2015) introduced attention into the RNN encoder in the QA setting. From then on, researchers have tried many kinds of ways to improve the attention mechanism on QA, like Yin et al. (2015); dos Santos et al. (2016); Wang et al. (2016). Wang et al. (2016) made a very successful attempt at doing impatient inner attention instead of the traditional outer attention over the hidden states of the sentences. They claim that this can make use of both the local word/phrase information and the sentence information. Wang and Jiang (2016) and Wang et al. (2017) apply a compare and aggregate framework on AS, and compare various ways to compute similarities between question and answer.

## 3    Our Approach

As is described in Yang et al. (2015), when they construct the WikiQA dataset, they first ask the annotators to decide whether the retrieved paragraph can answer the ques-tion. If so, the annotator is further asked to select which of the sentences can answer the question individually. Otherwise, each of the sentences in the paragraph is marked as *No*. Based on this observation, we assume that the overall information of the paragraph can be of help to predict the answer. Therefore, we propose our HGRNT model that aims to take the context information into consideration when calculating the confidence score of each candidate sentence.

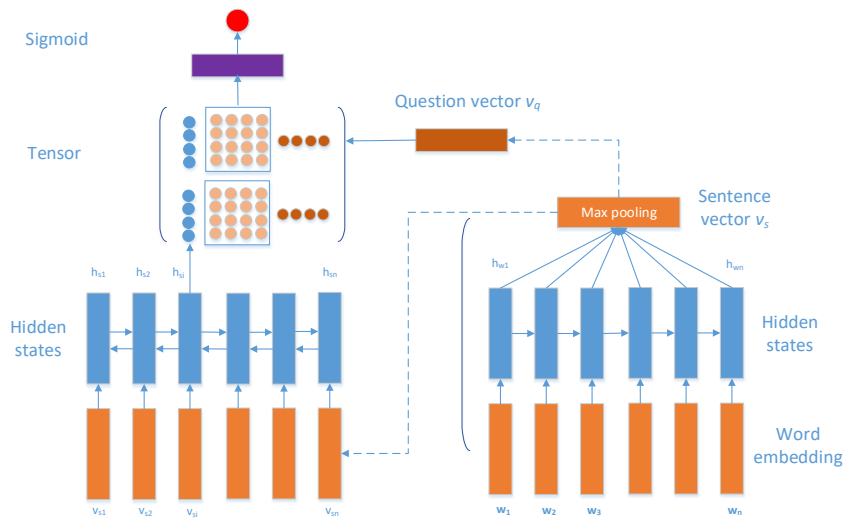

**Fig. 1.** Hierarchical Gated Recurrent Neural Tensor model for Answer Triggering problem

### 3.1 Hierarchical Gated Recurrent Neural Tensor model

Our approach is depicted in **Fig. 1**, we first encode the question sentence into a fixed length vector $v_q$ with the simple Gated Recurrent Neural Network (GRNN) (Cho et al., 2014). Then we encode answer sentences into vectors $v_s$ with another encoder. Different strategies of this answer sentence encoder can be applied. We will show the results of some models that have achieved state-of-the-art results on the AS problem in the next subsection.[1] The objectives of these models are very similar to our task except that they focus on the relative ranking scores of the sentences. In the bottom right part of **Fig. 1**, we present the encoder that gives the best result. Both the question encoder and the answer encoder are GRNN with max pooling. The dashed line in **Fig. 1** between max-pooling layer and $v_s$ or $v_q$ indicates that there is no transformation between these two parts.

After we get the vectors of the candidate sentences $v_s$, we go over the vector of each sentence in the paragraph with bidirectional gated recurrent neural networks

---

[1] We re-implement the model as the paper described, but we were not able to get as good as the original MRR and MAP result they claim. But this is not the focus of our paper

(BiGRNN), which lets the context information flow between answer sentences. Each sentence vector is treated as one time step in the BiGRNN. We denote the hidden states of the BiGRNN as $hs$, which capture the context information. We use BiGRNN because we think that context from both directions are important, and the gate mechanism can filter out the irrelevant information.

As is testified in Qiu and Huang (2015), neural tensor network is very effective in modelling the similarity between two sentences. After we get the answer sentence representation $hs$ produced by BiGRNN, we connect $hs$ with the question vector $v_q$ by a neural tensor layer as is shown in the top left part of **Fig. 1**, so that the deep interactions between the question and candidate sentences can be captured. The tensor layer can be calculated with Equation 1, where $v_q$ is the vector of the question, $hs$ is the hidden states of the candidate sentence $s$ produced by the BiGRNN, $f$ is a non-linear function, like *sigmoid*.

$$\mathrm{T}(q, a) = f\left(v_q M^{[1:r]} h_a\right) \qquad (1)$$

At last, we add a logistic regression layer to the model, which gives a confidence score of each sentence. The loss function is then set to be the negative log-likelihood between the score given by the logistic regression layer and the gold label (0 or 1) for each sentence in the paragraph. We set a threshold to decide whether to take the sentence with the highest score as the final answer. If the highest score is below the threshold, we reject all the sentences. Otherwise, we take the most probable sentence as the correct answer.

### 3.2    Sentence Encoder

The encoder of candidate sentences can be of various structures, which is not the focus of our paper. Here we list the ones we applied.

- Gated RNN: As is shown in the bottom right of **Fig. 1**, we use GRNN to go over each word embedding in the sentence, then max pooling is applied over the sentence length. The parameters of both candidate sentences and questions are shared.
- IARNN-Gate (Wang et al., 2016): This model is very similar to the GRNN model except that the question vector is first calculated and then is added to compute the gates of the answers. The details can be found in the original paper.
- Compare Aggregate model[2]: This model first performs word-level (context-level) matching, followed by aggregation using either CNN or RNN.

## 4    Experiment

In this paper, we conduct experiments on the WikiQA data. This data has already been split into train (70%), dev (10%) and test (20%) data. There are 3,047 questions in total, only 1,473 of which have answers. Each question is attached with a set of candidate sentences in a Wikipedia article.

---

[2] This kind of model is some what sophistecated, so we can only give a brief description. Please refer to Wang and Jiang (2016) and Wang et al. (2017) for detail.

All the hyper-parameters are tuned on the dev set. The word embeddings are pre-trained on the WikiQA corpus without fine-tuning by word2vec (Mikolov et al., 2013) tool-kit. We do our experiments using Tensorflow package (Abadi et al., 2015). The parameters in the model are all trained with Adam stochastic optimization method (Kingma and Ba, 2014). We use GRNN as the aggregate part for the Compare Aggregate model.

**Table 1.** Results compared with (Yang et al., 2015), IARNN-Gate (Wang et al., 2016), Compare and Aggregate (Wang and Jiang, 2016)

| Model | Prec | Rec | F |
|---|---|---|---|
| Yang et al. (2015) | 27.96 | 37.86 | 32.17 |
| IARNN - Gate | 25.94 | 42.39 | 32.19 |
| + context & tensor | 36.82 | 44.86 | 40.45 |
| compare aggregate | 27.64 | 39.92 | 32.65 |
| + context & tensor | 29.71 | 50.62 | 37.44 |
| GRNN | 38.03 | 25.51 | 30.54 |
| + context & tensor | 40.91 | 44.44 | 42.6 |

**Table 2.** Effect of adding context information and tensor

| Model | Prec | Rec | F |
|---|---|---|---|
| GRNN | 38.03 | 25.51 | 30.54 |
| + tensor | 39.36 | 30.45 | 34.34 |
| + context | 37.55 | 42.80 | 39.99 |
| + context & tensor | 40.91 | 44.44 | 42.6 |

### 4.1 Compare with baselines

From **Table 1** we can see, all the baseline models, even with state of the art MRR and MAP, get rather low F values which is the concern of our task. However, when these models are incorporated into our HGRNT framework, all of their F values are increased by a big margin. We think that this is because these models are only good at comparing the relative rank of sentences, but short at the ability to decide whether to accept the most probable sentence to be the answer. Context information becomes important in this situation. Additionally, these complicated models perform worse than our simple HGRNT, perhaps because they are affected by the scale of the corpus. Since the number of training samples is far from enough for such complicated neural models.

### 4.2 Effect of Context information

In this subsection, we analyze the effect of adding context information and tensor. As is shown in **Table 2**, the original GRNN model gives a poor result, with an F score of 30.54. However, both tensor network and context information give big improvements over the basic model. It is also worth noticing that the context information gives a significant gain in recall. This observation is consistent both with (30.45 ~ 44.44) and without (25.51 ~ 42.8) the tensor layer. We think that this is because with the help of context information our model can get hold of an overall idea about the whole re-trieved paragraph. In the next subsection, we will give an example of how this global information facilitates the predicting of individual sentences.

### 4.3 Case Study

In this subsection, we make a detailed analysis on two examples. We choose the neural tensor model without context information as the baseline, so that the effect of context information can be highlighted.

1. **Question:** *what is korean money called*
   **Candidates:** ① *the won (sign: ; code: krw) is the currency of south korea.* ② *a single won is divided into 100 jeon, the monetary subunit.* ③ *the jeon is no longer used for everyday transactions, and appears only in foreign exchange rates.*

2. **Question:** *where to write to mother angelica*
   **Candidates:** ① *mother mary angelica of the annunciation, pcpa (born rita antoinette rizzo on april 20, 1923) is an american franciscan nun best known as a television personality and the founder of the eternal word television network.* ② *in 1944, she entered the poor clares of perpetual adoration, a franciscan religious order for women, as a postulant, and a year later she was admitted to the order as a novice.* ③ *she went on to find a new house for the order in 1962 in irondale, alabama, where the ewtn is headquartered, and in 1996 she initiated the building of the shrine of the most blessed sacrament and our lady of the angels monastery in hanceville, alabama.* ④ *mother angelica hosted shows on ewtn until she suffered a stroke in 2001.* ⑤ *she is a recipient of the pro ecclesia et pontifice award granted by pope benedict xvi and lives in the cloistered monastery in hanceville.*

**Table 3.** Scores given by the Gated recurrent tensor model and HGRNT in example 1

| Id | golden label | Tensor | HGRNT |
|----|--------------|--------|-------|
| 1  | 1            | 0.2432 | 0.4924 |
| 2  | 0            | 0.0622 | 0.1362 |
| 3  | 0            | 0.0588 | 0.0073 |

**Table 4.** Scores given by the Gated recurrent tensor model and HGRNT in example 2

| Id | Golden label | Tensor | HGRNT |
|---|---|---|---|
| 1 | 0 | 0.3045 | 0.0237 |
| 2 | 0 | 0.0243 | 0.0132 |
| 3 | 0 | 0.0846 | 0.0588 |
| 4 | 0 | 0.0104 | 0.0075 |
| 5 | 0 | 0.1 | 0.0183 |

From **Table 3** we can see that in **Example 1**, although the relative rank of both models are the same, our HGRNT model gives a higher score on the first sentence, which is exactly the correct answer. In the example, we can observe that these three candidate sentences are all about Korean currency. The first sentence points out the answer, while the second sentence confirms the fact to be true by further dictating the relation between *won* and *jeon*, which are two Korean monetary subunits. This example shows that the context information can help predicting the score of individual sentences. It also explains why the recall rate is improved by a big margin when context information is added, as shown in **Table 2**. Additionally, from this example we can see that the models giving the same MRR and MAP may differ in F value, which explains why the state of the art models on answer selection task don't work well in our task.

In **Example 2** we can see that both models give very low scores on the second to the last sentences. The difference is that our HGRNT model makes the right decision by giving a rather low score on the first sentence. We think this is because the hierarchical structure can capture the context information and detect that the whole paragraph doesn't contain information about '*writing*' in the question.

## 5    Conclusion

In this paper, we employ a Hierarchical gated recurrent neural tensor model to deal with the answer triggering problem, which introduces the context information into our model. Our experiment result surpasses the baseline by over 10 %.

In the future we hope to develop a more sensible method to judge whether to accept a candidate sentence instead of setting a strict threshold. Additionally, the Wiki-iQA corpus is too small for sophisticated models to work, and we hope to make use of abundant unlabeled raw data to help resolve this problem.

**Acknowledgement.** This work is supported by the National Key Basic Research Program of China (2014CB340504), the National Natural Science Foundation of China (61371129, 61572245).

## References

1. Mart´ın Abadi, Ashish Agarwal, Paul Barham, Eugene Brevdo, Zhifeng Chen, Craig Citro, Greg S. Corrado, Andy Davis, Jeffrey Dean, Matthieu Devin, Sanjay Ghemawat, Ian

Goodfellow, Andrew Harp, Geoffrey Irving, Michael Isard, Yangqing Jia, Rafal Jozefowicz, Lukasz Kaiser, Manjunath Kudlur, Josh Levenberg, Dan Man´e, Rajat Monga, Sherry Moore, Derek Murray, Chris Olah, Mike Schuster, Jonathon Shlens, Benoit Steiner, Ilya Sutskever, Kunal Talwar, Paul Tucker, Vincent Vanhoucke, Vijay Vasudevan, Fernanda Vi´egas, Oriol Vinyals, Pete Warden, Martin Wattenberg, Martin Wicke, Yuan Yu, and Xiaoqiang Zheng. 2015. TensorFlow: Large-scale machine learning on heterogeneous systems. Software available from tensorflow.org. http://tensorflow.org/.

2. Dzmitry Bahdanau, Kyunghyun Cho, and Yoshua Bengio. 2014. Neural machine translation by jointly learning to align and translate. arXiv preprint arXiv:1409.0473 .

3. Kyunghyun Cho, Bart Van Merri¨enboer, Caglar Gulcehre, Dzmitry Bahdanau, Fethi Bougares, Holger Schwenk, and Yoshua Bengio. 2014. Learning phrase representations using rnn encoder-decoder for statistical machine translation. arXiv preprint arXiv:1406.1078 .

4. Cıcero Nogueira dos Santos, Ming Tan, Bing Xiang, and Bowen Zhou. 2016. Attentive pooling networks. CoRR, abs/1602.03609 .

5. Minwei Feng, Bing Xiang, Michael R Glass, Lidan Wang, and Bowen Zhou. 2015. Applying deep learning to answer selection: A study and an open task. In Automatic Speech Recognition and Understanding (ASRU), 2015 IEEE Workshop on. IEEE, pages 813–820.

6. Michael Heilman and Noah A Smith. 2010. Tree edit models for recognizing textual entailments, paraphrases, and answers to questions. In Human Language Technologies: The 2010 Annual Conference of the North American Chapter of the Association for Computational Linguistics. Association for Computational Linguistics, pages 1011–1019.

7. Karl Moritz Hermann, Tomas Kocisky, Edward Grefenstette, Lasse Espeholt,Will Kay, Mustafa Suleyman, and Phil Blunsom. 2015. Teaching machines to read and comprehend. In Advances in Neural Information Processing Systems. pages 1693–1701.

8. Diederik Kingma and Jimmy Ba. 2014. Adam: A method for stochastic optimization. arXiv preprint arXiv:1412.6980 .

9. Tomas Mikolov, Ilya Sutskever, Kai Chen, Greg S Corrado, and Jeff Dean. 2013. Distributed representations of words and phrases and their compositionality. In Advances in neural information processing systems. pages 3111–3119.

10. Xipeng Qiu and Xuanjing Huang. 2015. Convolutional neural tensor network architecture for communitybased question answering. In IJCAI. pages 1305– 1311.

11. Aliaksei Severyn and Alessandro Moschitti. 2013. Automatic feature engineering for answer selection and extraction. In EMNLP. volume 13, pages 458–467.

12. Ming Tan, Cicero dos Santos, Bing Xiang, and Bowen Zhou. 2015. Lstm-based deep learning models for non-factoid answer selection. arXiv preprint arXiv:1511.04108 .

13. Bingning Wang, Kang Liu, and Jun Zhao. 2016. Inner attention based recurrent neural networks for answer selection. In The Annual Meeting of the Association for Computational Linguistics.

14. Di Wang and Eric Nyberg. 2015. A long short-term memory model for answer sentence selection in question answering. In ACL (2). pages 707–712.

15. Mengqiu Wang, Noah A Smith, and Teruko Mitamura. 2007. What is the jeopardy model? a quasisynchronous grammar for qa. In EMNLP-CoNLL. volume 7, pages 22–32.

16. Shuohang Wang and Jing Jiang. 2016. A compare aggregate model for matching text sequences. arXiv preprint arXiv:1611.01747 .

17. Zhiguo Wang, Wael Hamza, and Radu Florian. 2017. Bilateral multi-perspective matching for natural language sentences. arXiv preprint arXiv:1702.03814

18. Yi Yang, Wen-tau Yih, and Christopher Meek. 2015. Wikiqa: A challenge dataset for open-domain question answering. In EMNLP. Citeseer, pages 2013– 2018.

19. Xuchen Yao, Benjamin Van Durme, Chris Callison- Burch, and Peter Clark. 2013. Answer extraction as sequence tagging with tree edit distance. In HLTNAACL. Citeseer, pages 858–867.

20. Wenpeng Yin, Hinrich Sch¨utze, Bing Xiang, and Bowen Zhou. 2015. Abcnn: Attention-based convolutional neural network for modeling sentence pairs. arXiv preprint arXiv:1512.05193 .